\documentclass[11pt]{article}

\usepackage[a4paper,margin=1in]{geometry}
\usepackage[T1]{fontenc}
\usepackage[utf8]{inputenc}
\usepackage{lmodern}
\usepackage{microtype}
\usepackage{amsmath,amssymb}
\usepackage{graphicx}
\usepackage{booktabs}
\usepackage{multirow}
\usepackage{array}
\usepackage{xcolor}
\usepackage{hyperref}
\usepackage{enumitem}
\usepackage{caption}
\usepackage{subcaption}
\usepackage{float}
\usepackage{tikz}
\usepackage{url}

\hypersetup{
    colorlinks=true,
    linkcolor=blue,
    citecolor=blue,
    urlcolor=blue
}

\title{Multimodal Object Detection Under Sparse Forest-Canopy Occlusion}

\author{
    Nitik Jain~\thanks{Robotics \& AI, Johns Hopkins University, USA. 
    This research was carried out entirely while the author was with the Department of Aerospace Engineering, IIT Kanpur.}
    \and
    Mangal Kothari~\thanks{Senior Principal  Flight Control Engineer, ADASI, EDGE Group, Abu Dhabi, UAE. 
    This research was carried out entirely while the author was with the Department of Aerospace Engineering, IIT Kanpur.}
}

\date{}

\begin{document}
\maketitle

\begin{abstract}
Reliable detection of humans beneath forest canopy remains a difficult remote-sensing challenge due to sparse, structured, and viewpoint-dependent occlusion. This paper presents a multimodal proof-of-concept pipeline that integrates three complementary approaches: (i) experimental evaluation of LiDAR returns through vegetation to assess the feasibility of active sensing, (ii) visible--thermal image fusion using a multi-scale transform and sparse-representation framework to enhance human saliency, and (iii) synthetic-aperture image formation via Airborne Optical Sectioning (AOS) to suppress canopy clutter. A YOLOv5 detector is fine-tuned on the Teledyne FLIR thermal dataset and evaluated on thermal and fused imagery. Results show that the tested terrestrial LiDAR configuration provides limited penetration for object-level detection, while visible--thermal fusion improves target visibility in low-contrast scenes and AOS enhances ground-plane detection in synthetic forest imagery. The fine-tuned YOLOv5 achieves a mean average precision of $\sim$0.83 on the top three FLIR classes. These findings establish an initial baseline for UAV-deployable search-and-rescue and surveillance systems operating in forested environments, and motivate future work on dedicated forest datasets and real-time multimodal integration.
\end{abstract}

\noindent\textbf{Keywords:} forest occlusion; object detection; thermal imaging; visible--infrared fusion; Airborne Optical Sectioning; synthetic aperture imaging; YOLOv5; UAV search and rescue.

\section{Introduction}

Occlusion due to vegetation is a fundamental challenge in remote sensing for applications such as search and rescue, border surveillance, wildlife monitoring, and disaster response. Forest canopies reduce direct line-of-sight visibility, scatter and attenuate optical signatures, and make object detection strongly dependent on viewpoint, sensor modality, illumination, and environmental conditions. In search-and-rescue scenarios, rapid detection of humans beneath sparse or partial canopy can directly influence operational response time and mission effectiveness. Similarly, in surveillance and security applications, autonomous detection from unmanned aerial vehicles (UAVs) can reduce human workload while enabling persistent wide-area monitoring.

Recent advances in computer vision, embedded AI, and UAV perception systems have enabled significant progress in autonomous object detection and scene understanding \cite{survey}. In particular, convolutional neural networks (CNNs) and single-stage object detectors such as the YOLO family of models have demonstrated real-time detection capability with favorable trade-offs between computational cost and accuracy \cite{yolo,yolov3}. These methods have found increasing deployment in robotic perception, UAV sensing, autonomous navigation, and surveillance applications.

Parallel developments in multimodal sensing and computational imaging have also shown promise for perception in degraded or partially occluded environments. Visible--thermal image fusion combines structural information from RGB imagery with thermal saliency from infrared sensing \cite{vifb,mstsr}, while synthetic-aperture and light-field imaging techniques exploit viewpoint diversity to suppress occluders and enhance target visibility \cite{lightfield,aos}. However, robust perception underneath sparse forest-canopy occlusion remains a difficult open problem due to highly non-uniform vegetation structure, varying illumination conditions, thermal clutter, and strong viewpoint dependence.

In many practical UAV deployments, additional constraints arise from limited payload capacity, onboard computational resources, power consumption, and sensing cost. Consequently, there remains a need for lightweight and deployable multimodal perception architectures that can improve detection reliability under forest occlusion without relying on prohibitively expensive sensing systems.

The present work also builds upon the authors' prior efforts in lightweight robotic perception and embedded computer-vision systems, including real-time drone detection and localization using lightweight YOLO-based architectures \cite{sharma2021multidrone}, autonomous robotic navigation using low-cost sensing \cite{jain2022agv}, and CNN-assisted robotic perception and localization pipelines \cite{sinha2018cnnloc}. While these earlier works primarily focused on structured or semi-structured robotic environments, the present work extends the investigation toward perception under sparse forest-canopy occlusion, where visibility, sensing geometry, and environmental uncertainty become significantly more challenging.

Motivated by these considerations, this paper presents a proof-of-concept multimodal pipeline for object detection underneath sparse forest-canopy occlusion. The central hypothesis is that no single sensing modality is sufficient for reliable detection in such environments, and that complementary sensing and image formation techniques must instead be combined into an integrated perception framework. Accordingly, this work investigates: (i) experimental evaluation of LiDAR sensing through vegetation, (ii) visible--thermal image fusion using multi-scale transform and sparse-representation techniques, (iii) synthetic-aperture image formation using Airborne Optical Sectioning (AOS), and (iv) lightweight CNN-based object detection using YOLOv5 trained on thermal imagery. The objective is not to claim a fully operational deployment-ready system, but rather to establish and experimentally evaluate a coherent baseline architecture that can guide future UAV-based perception systems operating in forested environments. The main contributions are:
\begin{enumerate}

\item A multimodal proof-of-concept architecture for object detection under sparse forest-canopy occlusion, combining active sensing, passive visible--thermal imaging, synthetic-aperture image formation, and CNN-based detection.

\item Experimental evidence that, for the tested FARO terrestrial laser scanner configuration and vegetation scenarios, LiDAR returns through canopy are insufficiently reliable for direct object-level detection beneath vegetation.

\item Implementation and evaluation of visible--thermal fusion using a multi-scale transform and sparse-representation (MST--SR) framework to enhance human/object saliency in low-contrast and partially occluded scenes.

\item Application of Airborne Optical Sectioning (AOS) on synthetic forest data to demonstrate that synthetic-aperture integration can suppress vegetation occluders and improve visibility of objects at the forest floor.

\item Fine-tuning and evaluation of YOLOv5 on thermal imagery, achieving approximately 0.83 mAP on the top three classes in the Teledyne FLIR test set.

\end{enumerate}

\section{Problem Formulation and System Overview}
The problem considered in this work is the detection of humans or human-scale objects under sparse or partial forest-canopy occlusion. Let an observed scene contain a target object $O$ located near the ground plane and a set of vegetation occluders $\{V_i\}$ distributed between the sensor and the target. A sensor observation may be written abstractly as
\begin{equation}
    I_m = \mathcal{H}_m(O, V, \theta_m, \eta_m),
\end{equation}
where $I_m$ is the image or measurement from modality $m$, $\theta_m$ represents sensor pose and intrinsic parameters, and $\eta_m$ represents noise and environmental effects. The task is to estimate object class and location,
\begin{equation}
    \{c_k, b_k, s_k\}_{k=1}^{N} = \mathcal{D}(I_1,I_2,\ldots,I_M),
\end{equation}
where $c_k$ is the class label, $b_k$ is a bounding box, and $s_k$ is a confidence score. In practice, $\mathcal{D}(\cdot)$ is implemented by a convolutional neural network detector trained on thermal imagery and evaluated on fused or enhanced inputs.

\subsection*{Pipeline Overview}
\begin{figure}[H]
\centering
\includegraphics[width=0.95\textwidth]{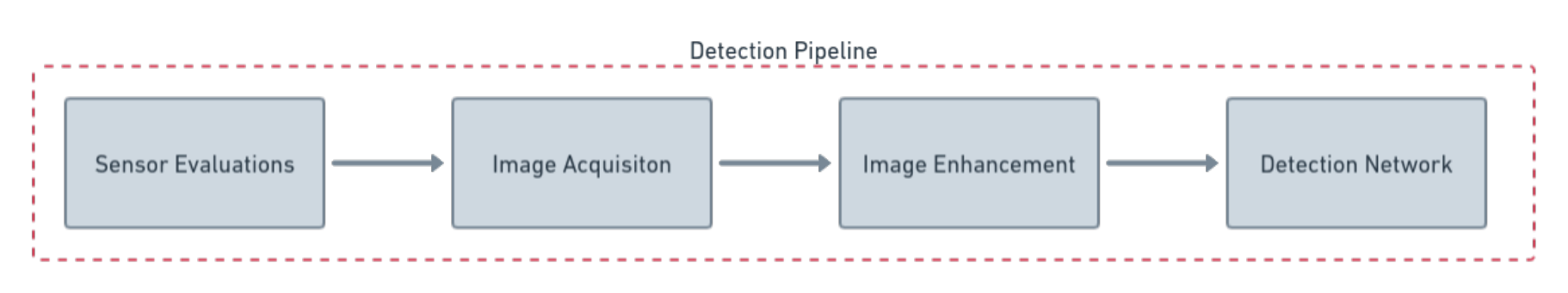}
\caption{Conceptual pipeline for multimodal object detection underneath sparse forest-canopy occlusion. The framework integrates active LiDAR sensing, passive RGB--thermal fusion, and synthetic-aperture imaging before CNN-based detection.}
\label{fig:pipeline}
\end{figure}

The proposed proof-of-concept pipeline (Fig.~\ref{fig:pipeline}) evaluates three complementary sensing and enhancement routes before passing the processed measurements to an object detector:

\begin{enumerate}
    \item \textbf{Active LiDAR sensing:} A terrestrial laser scanner is used to test whether point-cloud returns through sparse vegetation contain sufficient geometric information for object-level detection. This route probes the feasibility of foliage penetration and geometric reconstruction.
    \item \textbf{Passive RGB--thermal fusion:} Co-registered visible and thermal images are fused using a multi-scale transform and sparse-representation (MST-SR) framework. RGB imagery contributes structural detail and context, while thermal imagery provides saliency for humans and animals. The fusion stage aims to improve separability of targets from clutter in single-frame observations.
    \item \textbf{Synthetic-aperture imaging (AOS):} Multi-view imagery is integrated using Airborne Optical Sectioning, which reprojects and refocuses images on the ground plane. This suppresses inconsistent canopy occluders and enhances visibility of objects beneath sparse foliage. AOS exploits viewpoint diversity rather than per-frame fusion.
\end{enumerate}

\subsection*{Integration with Detection}
The enhanced measurements from these routes are passed to a YOLOv5 detector fine-tuned on thermal imagery. The detector outputs bounding boxes, class labels, and confidence scores. This modular design allows each sensing route to be evaluated independently while maintaining a unified downstream detection framework.

\section{Sensor Modalities and Data Sources}

\subsection{Active and Passive Sensors}
Remote sensing modalities can be broadly categorized into \emph{active} and \emph{passive} sensors. Active sensors such as LiDAR or radar transmit energy and measure the returned signal, providing geometric or penetrative information. Passive sensors such as RGB and thermal cameras measure naturally reflected or emitted energy, offering contextual and saliency cues. For UAV-compatible low-altitude search, the trade-off between payload, power, and sensing utility is critical: RGB and thermal cameras are lightweight and mature, while compact LiDAR units provide geometric context but at higher cost and complexity.

Table~\ref{tab:sensors} summarizes the qualitative comparison motivating the experimental choices. The values are relative and scenario-dependent; the intent is to identify practical sensing candidates for UAV deployment under sparse canopy rather than provide universal sensor rankings.

\begin{table}[H]
\centering
\caption{Qualitative comparison of candidate sensing modalities for low-altitude detection under forest-canopy occlusion.}
\label{tab:sensors}
\begin{tabular}{p{3cm}p{2cm}p{2.4cm}p{2.3cm}p{4.2cm}}
\toprule
\textbf{Sensor} & \textbf{Type} & \textbf{Relative cost} & \textbf{UAV deployment} & \textbf{Expected utility under sparse canopy} \\
\midrule
RGB camera & Passive & Low & High & Provides structural detail and texture; effective under good illumination and line-of-sight, but weak in shadows or dense canopy. \\
Thermal camera & Passive & Medium & High & Highlights humans/animals via heat contrast; performance degrades under dense occlusion or thermal clutter. \\
LiDAR & Active & Medium--High & Medium & Captures geometry and depth; penetration depends on wavelength, sampling density, canopy structure, and return strength. \\
SAR radar & Active & High & Low--Medium & Offers potential foliage penetration; however, system complexity, antenna size, and cost limit UAV integration. \\
Hyperspectral/multispectral & Passive & High & Medium & Useful for vegetation/material analysis; limited direct human-detection capability in forested settings. \\
\bottomrule
\end{tabular}
\end{table}

\subsection{Thermal and RGB Imaging}
Thermal cameras measure infrared radiation emitted by objects above absolute zero, making them attractive for human detection since people often exhibit strong contrast against natural backgrounds, especially in low-light or shadowed scenes. RGB imagery, by contrast, provides structural detail, texture, and contextual cues. The proof-of-concept experiments employed a CAT S60 smartphone equipped with an RGB camera and a FLIR Lepton thermal module. This configuration allowed co-registered acquisition of visible and thermal data at low cost, suitable for evaluating fusion strategies.

\begin{figure}[H]
\centering
\includegraphics[width=0.95\textwidth]{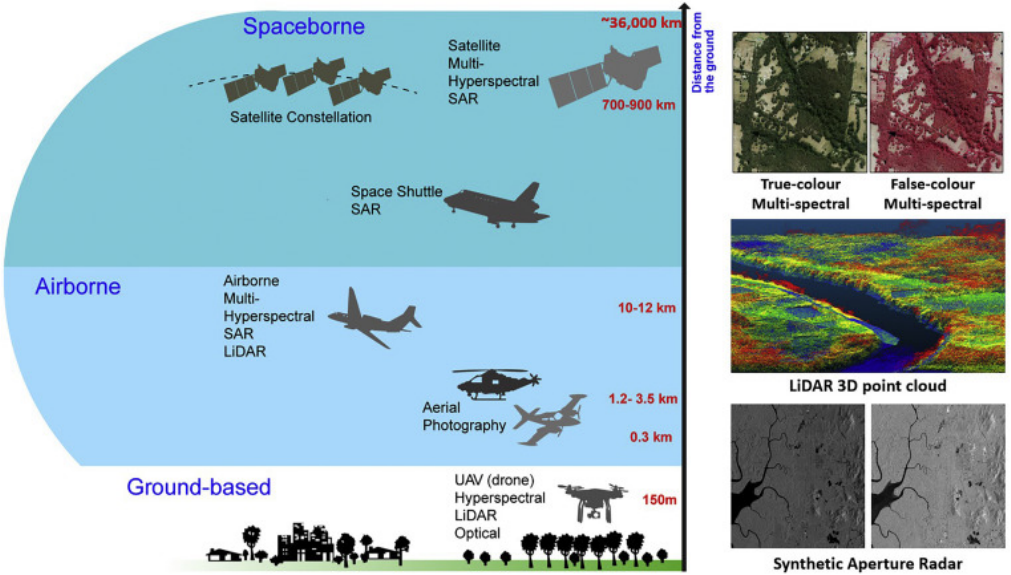}
\caption{Remote-sensing sensor classes and operational regimes considered when selecting UAV-compatible modalities for low-altitude forest search.}
\label{fig:remote_sensing}
\end{figure}

\begin{figure}[H]
\centering
\includegraphics[width=0.72\textwidth]{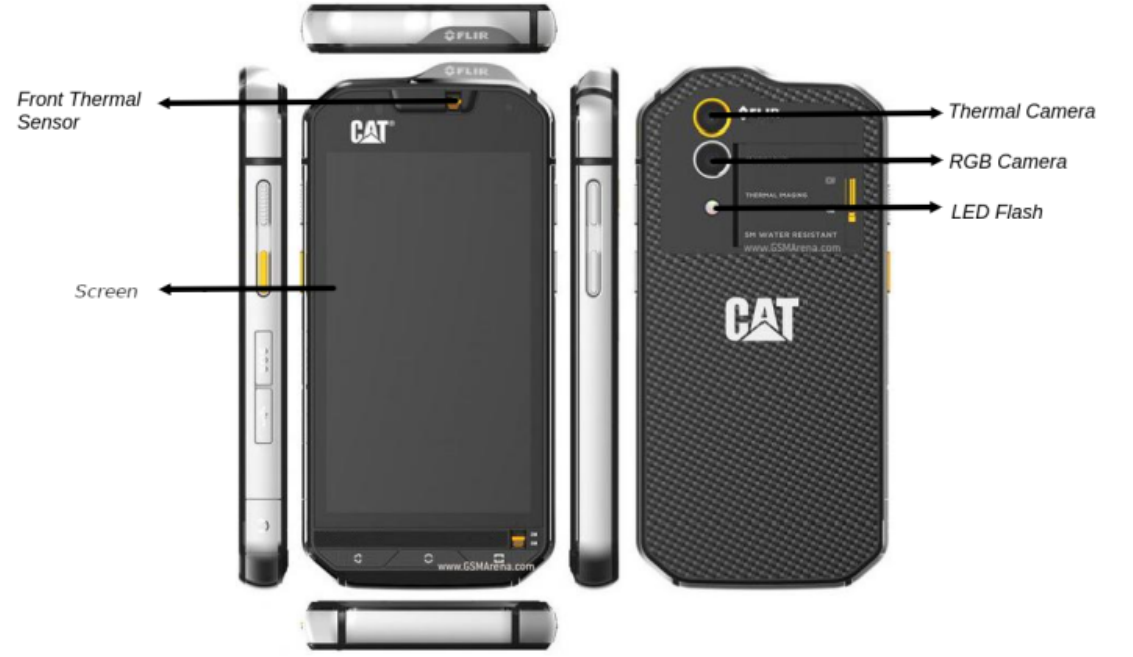}
\caption{CAT S60 smartphone used for proof-of-concept RGB and thermal image acquisition.}
\label{fig:cat_s60}
\end{figure}

\subsection{Datasets and Simulation Environment}
The detection component leverages the Teledyne FLIR ADAS thermal dataset for training and evaluation. This dataset provides paired thermal and visible-spectrum imagery with annotated objects, enabling study of thermal detection and visible--thermal fusion. Transfer learning from the COCO dataset supplies pretrained object-recognition knowledge, improving detector generalization.

To evaluate synthetic-aperture imaging, a controlled simulation environment was constructed. Procedurally generated trees and human targets were placed in a synthetic forest scene, and multiple camera views were rendered from diverse viewpoints. This multi-view dataset allowed systematic study of Airborne Optical Sectioning (AOS), isolating the benefits of viewpoint diversity in suppressing canopy occluders. While synthetic, the environment provided controlled conditions for algorithmic validation prior to real-world UAV deployment.

\section{Methodology}

\subsection{LiDAR Experiments}
The LiDAR experiments were designed to evaluate whether point-cloud returns contain sufficient target information through sparse vegetation to support object-level detection. A FARO Focus Premium 350 terrestrial laser scanner was mounted on a fixed tripod. A white thermocol rectangle was used as a high-reflectance surrogate target, creating a favorable case for detecting returns through vegetation. Two representative scenarios were tested:
\begin{enumerate}[leftmargin=*]
    \item \textbf{Ground-level occlusion:} target concealed behind $\sim$30~cm thick vegetation, representing sparse horizontal occlusion at human scale.
    \item \textbf{Elevated canopy view:} scanner positioned at a nearby elevated vantage point, representing a UAV-like nadir view through tree canopy.
\end{enumerate}

The scanner parameters are summarized in Table~\ref{tab:lidarparams}. The experiment intentionally used a static sensor setup and a high-reflectance target; therefore, failure to obtain useful returns in this configuration strongly indicates that direct LiDAR-based detection is difficult under the considered canopy conditions. These experiments serve as a negative baseline, narrowing the design space and motivating multimodal alternatives.

\begin{table}[H]
\centering
\caption{LiDAR scan parameters used in the proof-of-concept experiments.}
\label{tab:lidarparams}
\begin{tabular}{lcccc}
\toprule
\textbf{Experiment} & \textbf{Points} & \textbf{Duration} & \textbf{Resolution} & \textbf{Quality} \\
\midrule
Ground-level occlusion & 29.3 million & $\sim$9 min & 1/2 & 2x \\
Elevated canopy view & 9.8 million & $\sim$7 min & 1/2 & 3x \\
\bottomrule
\end{tabular}
\end{table}

\subsection{Visible--Thermal Fusion}
Visible--thermal fusion combines the structural detail of RGB imagery with thermal saliency from infrared imagery. This work employs the Multi-Scale Transform and Sparse Representation (MST--SR) framework, where source images are decomposed into multi-scale components and informative features are selectively combined. 

Let $I_v$ and $I_t$ denote visible and thermal images. The fusion operation is expressed as
\begin{equation}
    I_f = \mathcal{F}_{\mathrm{MST-SR}}(I_v,I_t),
\end{equation}
where $I_f$ is the fused image. Thermal signatures highlight humans and animals, while RGB texture preserves scene context and boundaries. For downstream detection, the fused image improves object--background separability without destroying spatial localization cues. This approach is computationally lightweight compared to dense 3D reconstruction, making it suitable for real-time UAV deployment.

\subsection{Airborne Optical Sectioning}
Airborne Optical Sectioning (AOS) is a synthetic-aperture imaging method based on light-field rendering. Instead of relying on a single viewpoint, a set of images is captured from multiple perspectives. These images are reprojected and integrated such that a chosen focal plane, typically the ground plane, is reinforced while out-of-focus occluders such as canopy leaves are blurred or suppressed.

Let $L(s,t,u,v)$ denote a light field parameterized by camera-plane coordinates $(s,t)$ and image-plane coordinates $(u,v)$. A synthetic refocused image at focal surface $F$ can be expressed as
\begin{equation}
    I_F(u,v) = \frac{1}{N}\sum_{i=1}^{N} W_i(u,v)\, I_i\big(\pi_i(F,u,v)\big),
\end{equation}
where $I_i$ is the $i$th camera image, $\pi_i$ is the projection/warping operation corresponding to the chosen focal surface, $W_i$ is a visibility or weighting term, and $N$ is the number of views. When viewpoints are sufficiently diverse, vegetation inconsistent across views is attenuated by integration. This principle allows canopy clutter to be suppressed without requiring dense 3D reconstruction.

\begin{figure}[H]
\centering
\includegraphics[width=0.92\textwidth]{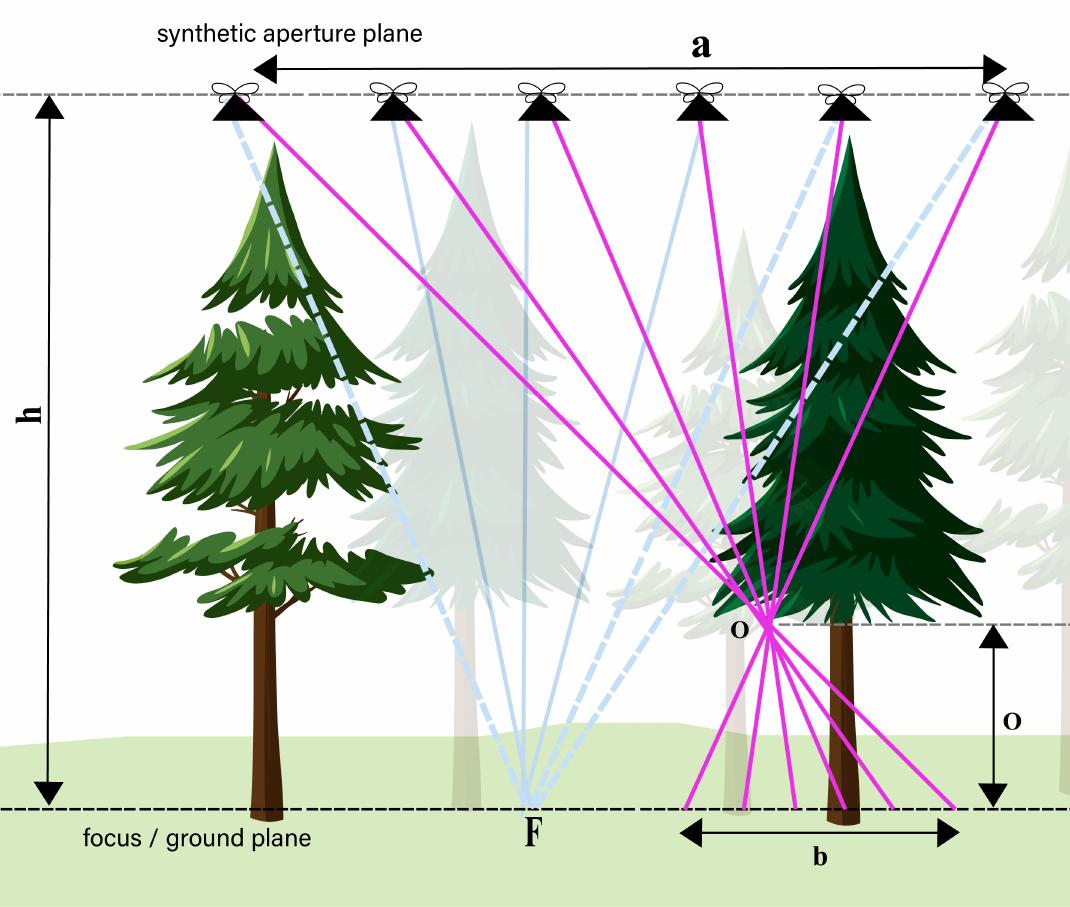}
\caption{Airborne Optical Sectioning principle: multiple images captured over a synthetic aperture are refocused on a selected ground-plane target while canopy occluders are defocused.}
\label{fig:aos_method}
\end{figure}

\subsection{YOLOv5 Detection}
The detection stage is based on YOLOv5, chosen for its favorable trade-off between accuracy, model size, and inference speed. Transfer learning is employed: the network is initialized from pretrained weights (COCO dataset) and fine-tuned on thermal imagery (Teledyne FLIR dataset). This leverages pretrained feature extraction while adapting to thermal signatures.

Detection outputs are evaluated using standard object-detection metrics: precision, recall, intersection-over-union (IoU), average precision (AP), and mean average precision (mAP). For a predicted bounding box $B_p$ and ground-truth bounding box $B_g$, IoU is defined as
\begin{equation}
    \mathrm{IoU}(B_p,B_g)=\frac{|B_p \cap B_g|}{|B_p \cup B_g|}.
\end{equation}
Average precision summarizes the precision--recall curve for each class, while mAP averages AP across classes. The fine-tuned YOLOv5 detector thus provides a lightweight, deployable inference engine suitable for embedded UAV platforms.

\section{Experimental Results}

\subsection{LiDAR Observations}
In the ground-level vegetation experiment, the point cloud was dense overall but did not provide reliable returns from the target behind leaves. Even when the target was partially visible optically, the returned point cloud lacked separable target structure. In the elevated canopy experiment, the scanner mapped the upper canopy and non-occluded objects, but returns below the canopy were too sparse to be useful for object-level detection.

The key inference is that adding this LiDAR configuration to the detection pipeline does not provide meaningful benefit for the considered sparse-canopy task. While LiDAR remains valuable for geometric mapping and terrain modeling, its utility for direct human detection beneath canopy is limited under the tested conditions. Alternative configurations (e.g., closer-range mounting, different wavelengths, or fusion with thermal/RGB imagery) may still offer promise, but as tested here, LiDAR does not solve the occlusion problem.

\begin{figure}[H]
\centering
\includegraphics[width=0.95\textwidth]{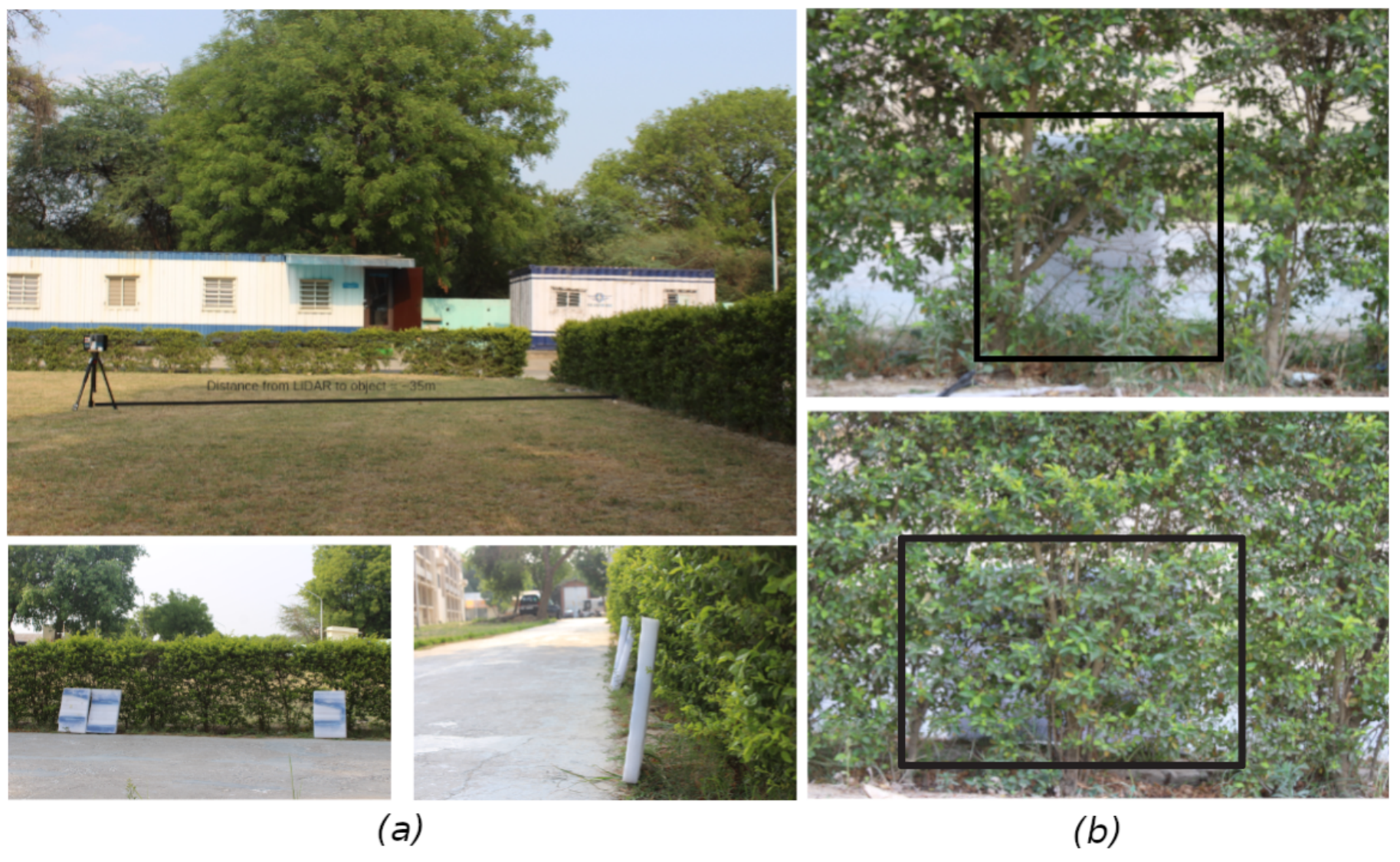}
\caption{LiDAR experiment with target partially concealed behind ground-level vegetation.}
\label{fig:lidar_exp1}
\end{figure}

\begin{figure}[H]
\centering
\includegraphics[width=0.95\textwidth]{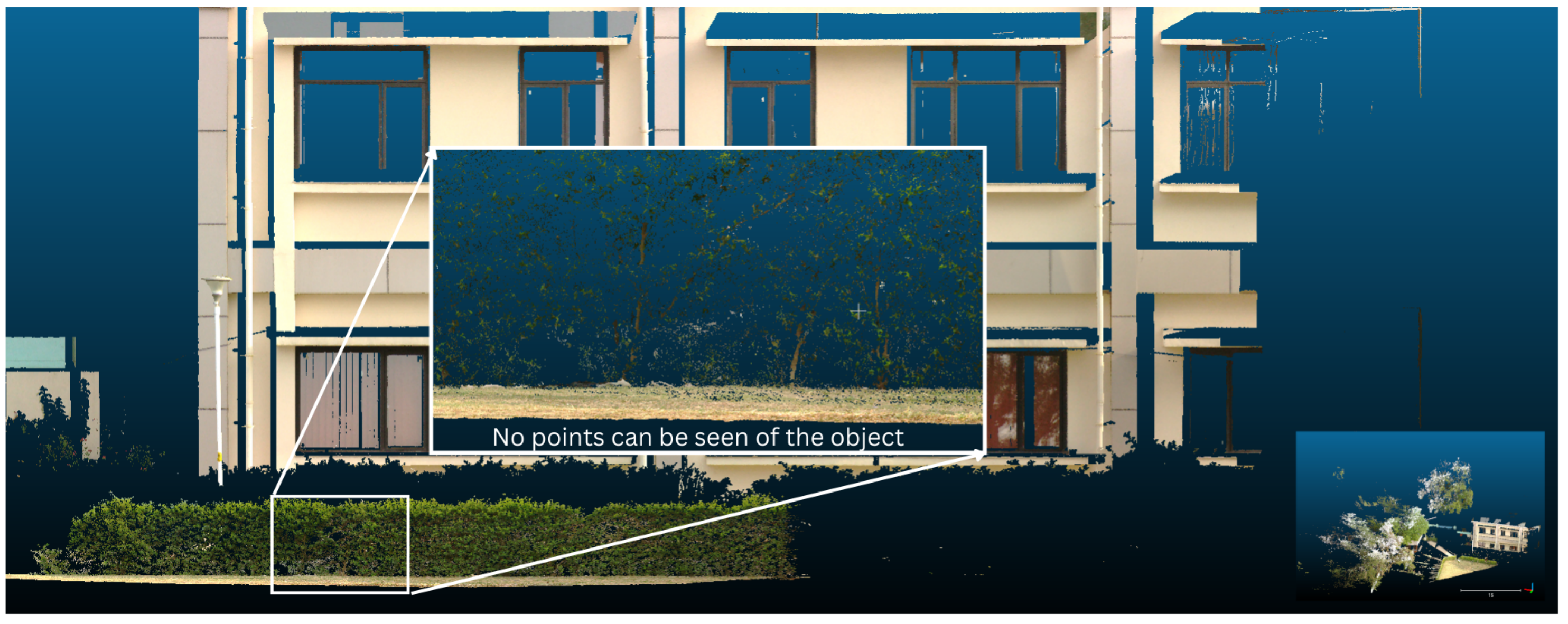}
\caption{Point-cloud observation from the first LiDAR experiment. The target structure is not reliably separable behind sparse vegetation.}
\label{fig:lidar_obs1}
\end{figure}

\begin{figure}[H]
\centering
\includegraphics[width=0.95\textwidth]{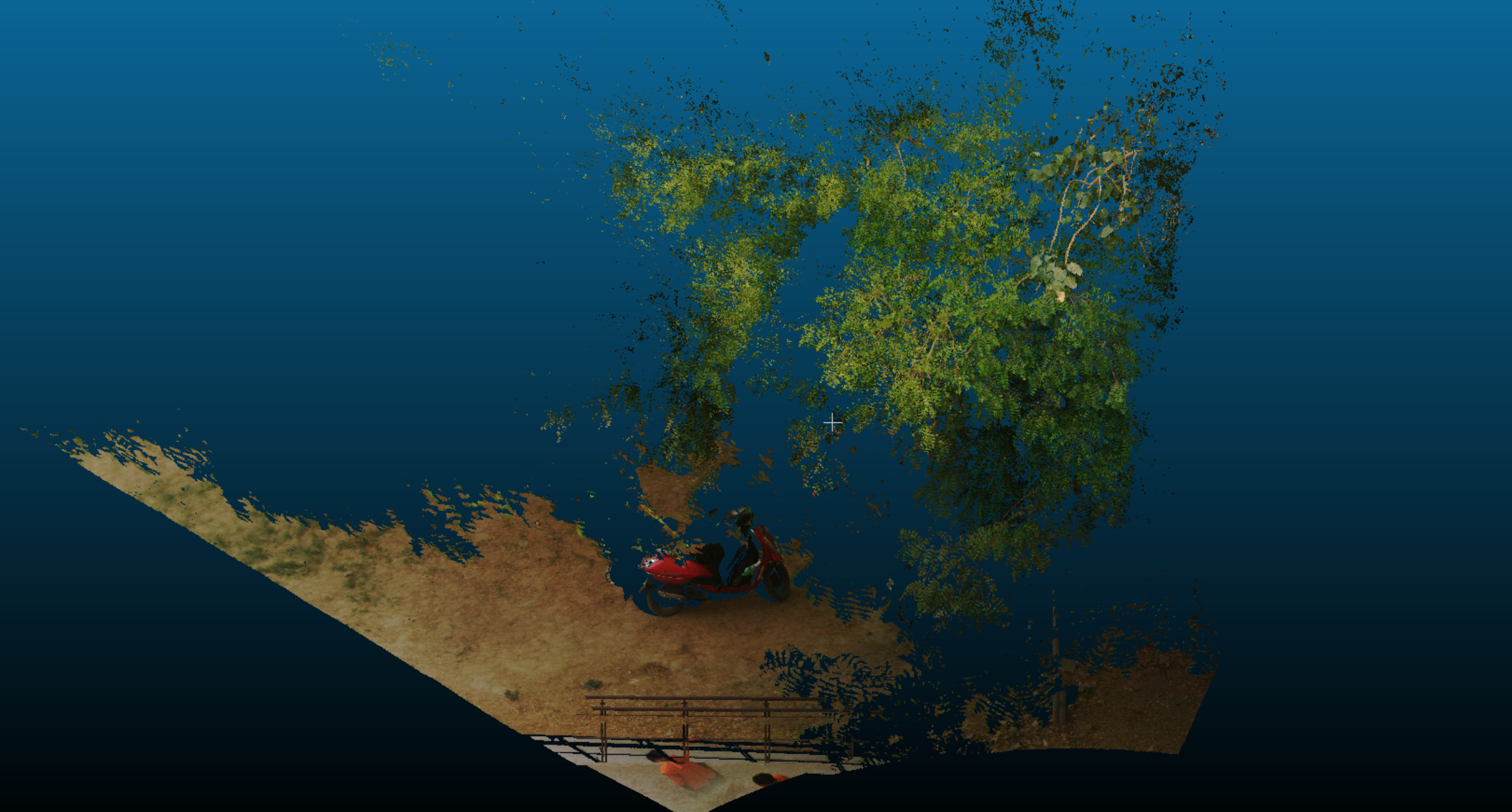}
\caption{Elevated/top-down LiDAR observation through canopy. The scanner primarily registers canopy and visible structure, with insufficient useful returns below the canopy.}
\label{fig:lidar_obs2}
\end{figure}

\subsection{Visible--Thermal Fusion Results}
The MST--SR fusion framework was evaluated on benchmark visible--infrared image pairs and self-acquired RGB/thermal scenes. Qualitatively, fusion improved visibility of humans in shadows, low-contrast backgrounds, and partial vegetation occlusion. Thermal imagery provided strong target saliency, while RGB imagery preserved object boundaries and scene context.

The most important practical outcome is that visible--thermal fusion appears to be a promising low-cost enhancement stage for UAV-based search. Unlike dense 3D reconstruction or semantic mapping, MST--SR fusion is computationally lightweight and suitable for real-time onboard implementation. This makes it an attractive candidate for embedded UAV pipelines where payload and compute budgets are constrained.

\begin{figure}[H]
\centering
\includegraphics[width=0.95\textwidth]{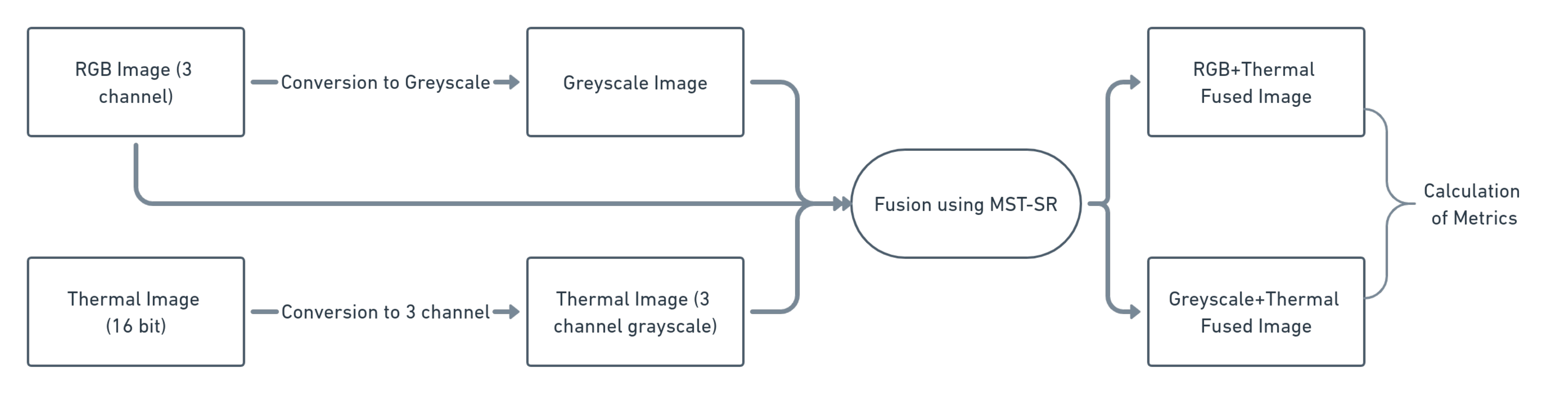}
\caption{MST--SR visible--thermal image-fusion processing flow.}
\label{fig:fusion_framework}
\end{figure}

\begin{figure}[H]
\centering
\includegraphics[width=0.95\textwidth]{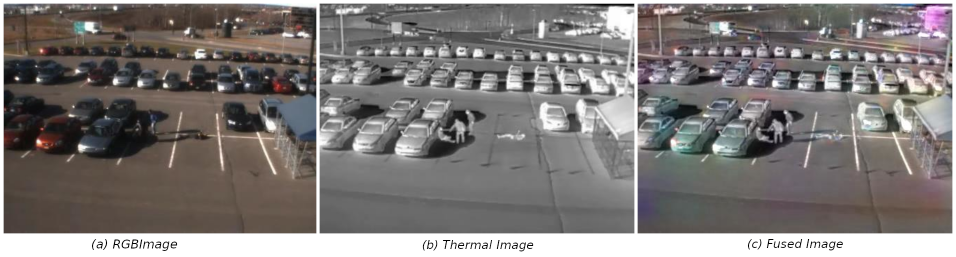}
\caption{Example visible--thermal fusion result on a benchmark scene containing people under challenging contrast and illumination.}
\label{fig:fusion_benchmark}
\end{figure}

\begin{figure}[H]
\centering
\includegraphics[width=0.95\textwidth]{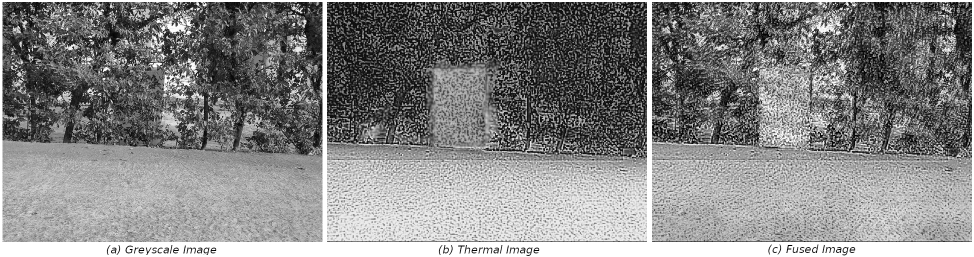}
\caption{Self-acquired RGB--thermal fusion example with object/human signature partially affected by vegetation occlusion.}
\label{fig:fusion_own}
\end{figure}

\subsection{YOLOv5 Thermal Detection}
YOLOv5 was fine-tuned on the Teledyne FLIR thermal dataset. The dataset is imbalanced, with cars dominating and fewer instances of pedestrians and bicycles. The trained detector achieved the best performance on the dominant classes. The reported mean average precision for the top three classes in the test set is approximately $0.83$. Table~\ref{tab:map} summarizes the results.

\begin{table}[H]
\centering
\caption{Summary of YOLOv5 detection performance on the FLIR thermal test set.}
\label{tab:map}
\begin{tabular}{lc}
\toprule
\textbf{Evaluation setting} & \textbf{Reported performance} \\
\midrule
Top three FLIR classes & mAP $\approx 0.83$ \\
Inference suitability & Low-latency, lightweight model suitable for embedded exploration \\
Primary limitation & Dataset imbalance and domain gap to forest scenes \\
\bottomrule
\end{tabular}
\end{table}

\begin{figure}[H]
\centering
\includegraphics[width=0.72\textwidth]{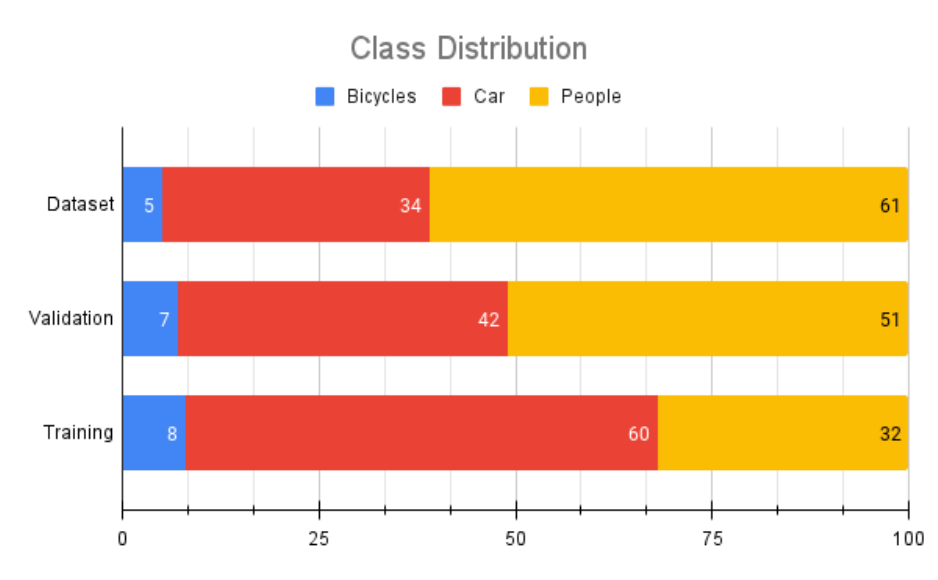}
\caption{Class distribution in the FLIR thermal dataset, showing dataset imbalance relevant to detector performance.}
\label{fig:flir_class_distribution}
\end{figure}

\begin{figure}[H]
\centering
\includegraphics[width=0.95\textwidth]{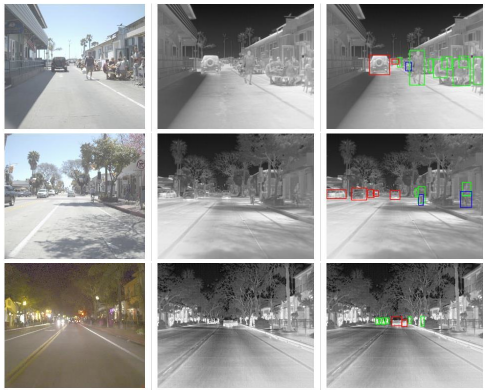}
\caption{YOLOv5 inference results on FLIR thermal test imagery. Bounding boxes show detected bicycles, cars, and people.}
\label{fig:yolo_results}
\end{figure}

Although encouraging, these results should not be interpreted as final forest-canopy detection accuracy. The training and test data are from automotive thermal scenes, whereas the target deployment environment is forested terrain. Thus, the results demonstrate transfer-learning feasibility and detector readiness, but a dedicated forest thermal dataset is required for operational validation.

\subsection{AOS on Synthetic Forest Data}
AOS was evaluated using synthetic forest imagery with human targets on the forest floor. Multiple images were rendered from different viewpoints and combined to form an integral image focused near the ground plane. The resulting image showed reduced vegetation clutter and improved visibility of partially occluded humans compared with individual views.

This experiment supports the central idea that viewpoint diversity can be exploited to suppress sparse occluders. However, the result was obtained from synthetic data with controlled camera poses and limited sensor noise. Real-world AOS will require accurate camera pose estimation, synchronization, image registration, and robustness to illumination variation, wind-driven vegetation motion, rolling-shutter effects, and thermal/RGB calibration errors. Despite these challenges, the synthetic evaluation demonstrates the potential of AOS as a complementary module in multimodal UAV perception pipelines.

\begin{figure}[H]
\centering
\includegraphics[width=0.95\textwidth]{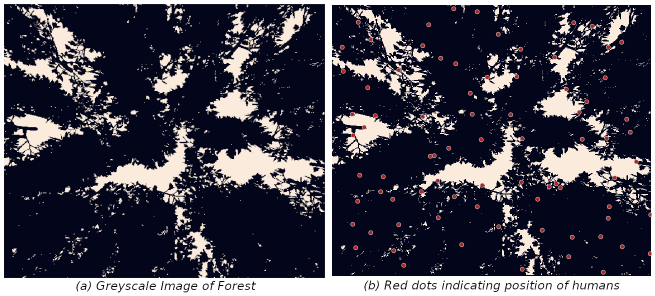}
\caption{Synthetic forest scene used for AOS evaluation. Human targets are placed on the forest floor under sparse canopy.}
\label{fig:synthetic_forest}
\end{figure}

\begin{figure}[H]
\centering
\begin{subfigure}{0.31\textwidth}\centering\includegraphics[width=\linewidth]{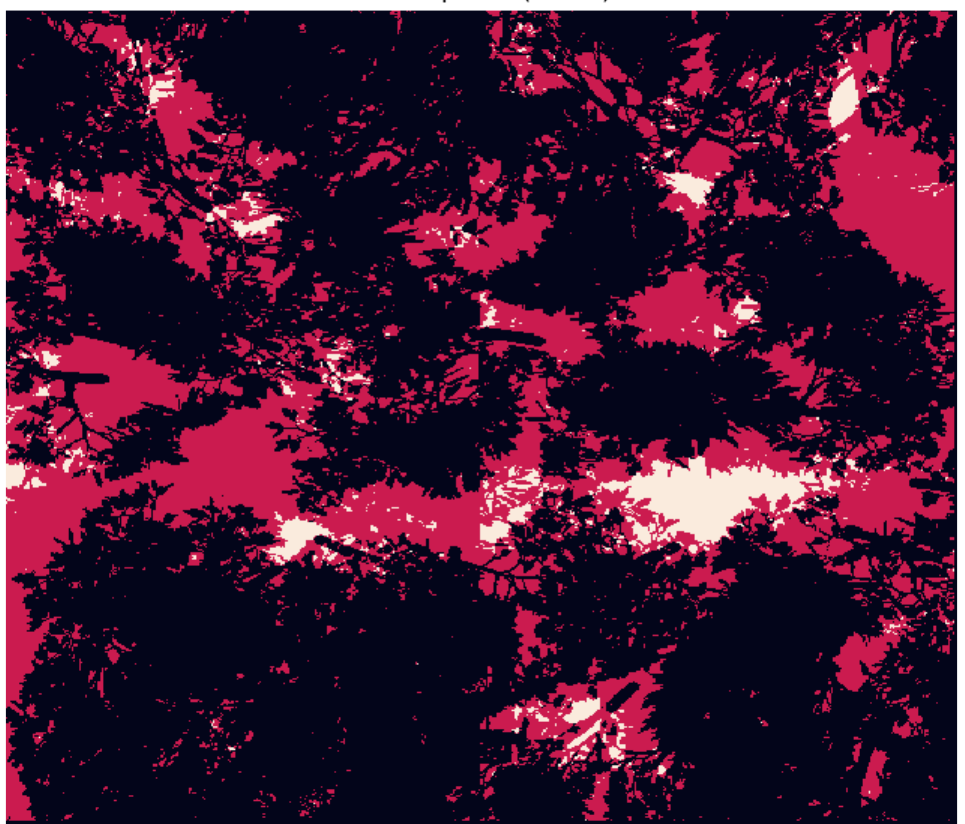}\caption{View 1}\end{subfigure}\hfill
\begin{subfigure}{0.31\textwidth}\centering\includegraphics[width=\linewidth]{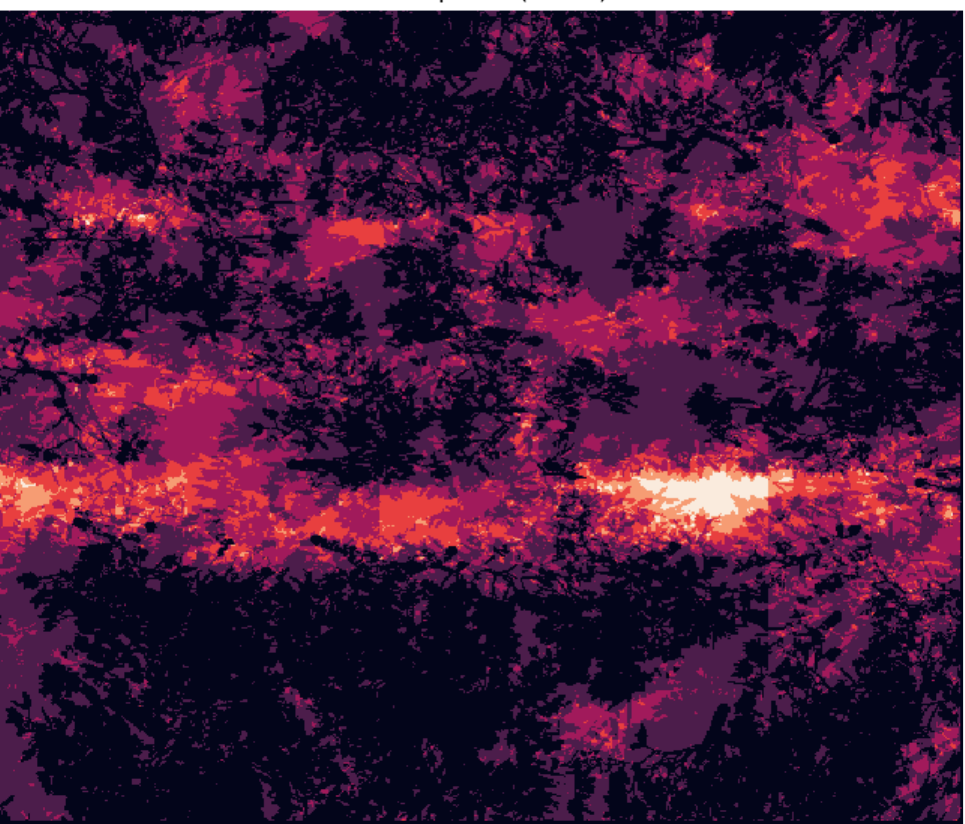}\caption{View 2}\end{subfigure}\hfill
\begin{subfigure}{0.31\textwidth}\centering\includegraphics[width=\linewidth]{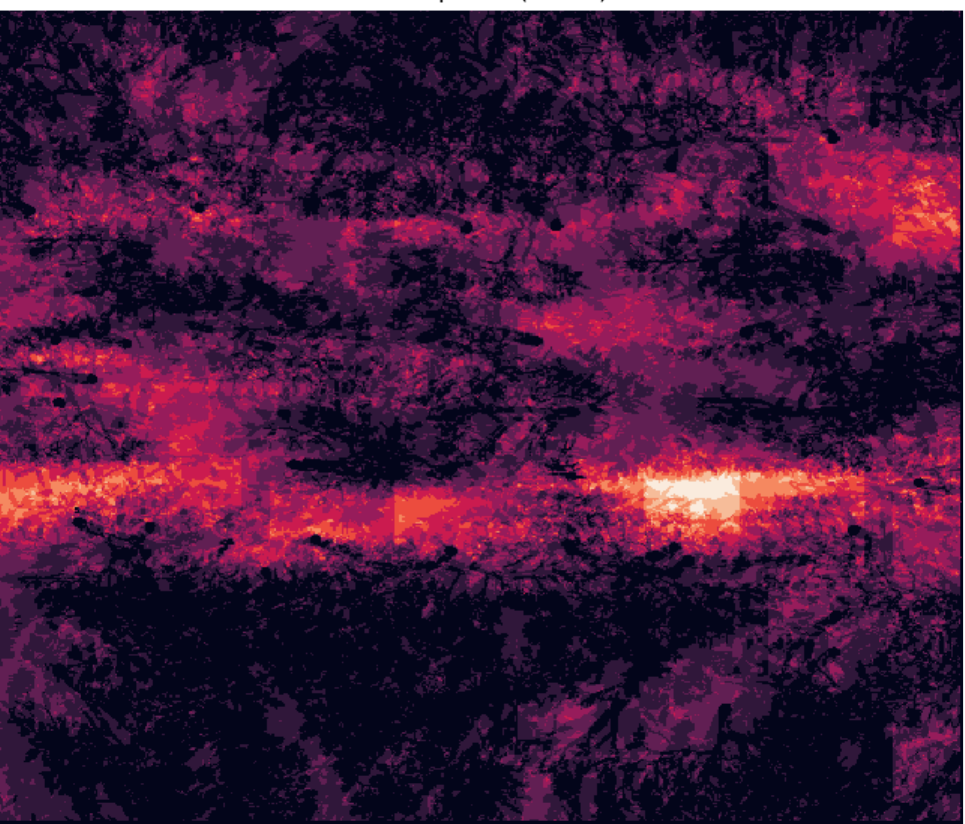}\caption{View 3}\end{subfigure}
\caption{Representative individual images from the synthetic multi-view sequence used for AOS. Each view contains different canopy occlusion patterns.}
\label{fig:aos_views}
\end{figure}

\begin{figure}[H]
\centering
\includegraphics[width=0.86\textwidth]{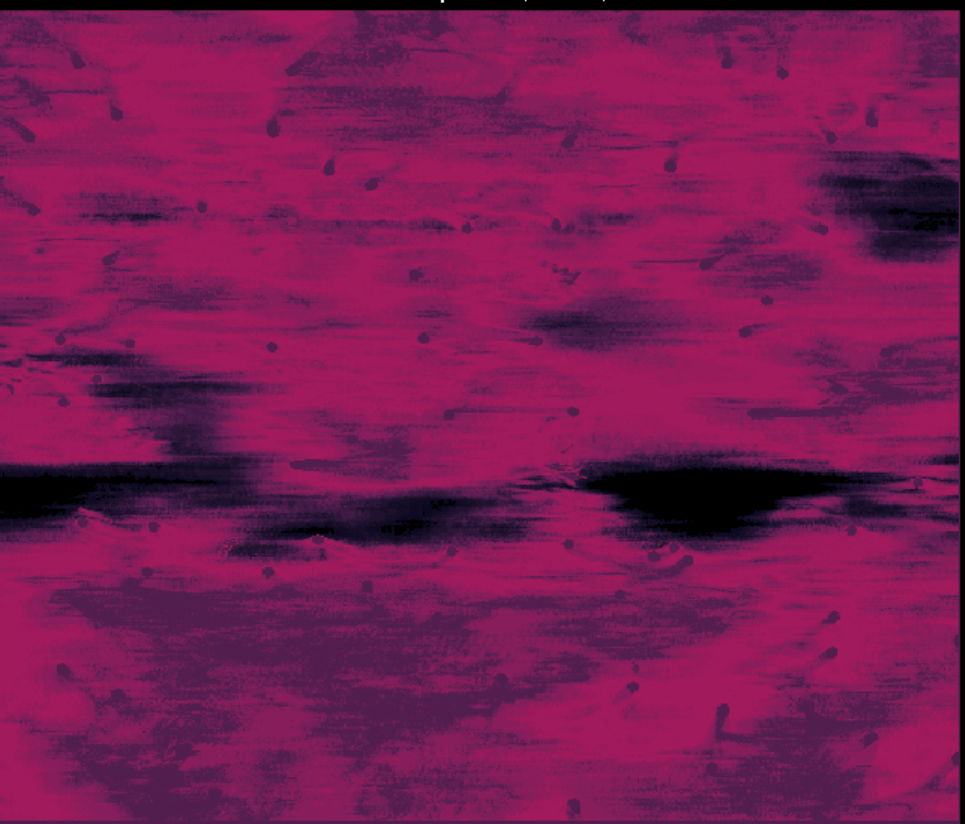}
\caption{Final AOS integral image produced by combining the synthetic multi-view sequence. Vegetation clutter is reduced and ground-plane target visibility is enhanced.}
\label{fig:aos_integral}
\end{figure}

\section{Discussion}
The experiments collectively highlight that the forest-occlusion detection problem must be treated as a \emph{multimodal perception challenge} rather than a single-sensor task. Each modality contributes distinct advantages and limitations:

\begin{itemize}[leftmargin=*]
    \item \textbf{LiDAR:} The negative results are valuable in narrowing the design space. Despite dense point clouds, sparse canopy penetration was insufficient for reliable human detection. This suggests that straightforward LiDAR integration is unlikely to yield operational benefit unless combined with alternative wavelengths, closer-range deployment, or geometric fusion with passive imagery.
    \item \textbf{Visible--thermal fusion:} Fusion improved per-frame saliency by combining structural detail with thermal contrast. Importantly, the MST--SR framework is computationally lightweight, making it suitable for UAV platforms with limited onboard resources. This positions fusion as a practical enhancement stage for real-time search-and-rescue.
    \item \textbf{Airborne Optical Sectioning (AOS):} Synthetic experiments demonstrated that viewpoint diversity can suppress canopy clutter and enhance ground-plane visibility. While promising, real-world deployment will require robust pose estimation, synchronization, and resilience to environmental dynamics such as wind-driven motion and illumination variation.
    \item \textbf{YOLOv5 detection:} The fine-tuned detector achieved encouraging performance ($\sim$0.83 mAP) on FLIR thermal classes, validating transfer learning feasibility. However, the dataset imbalance and domain gap to forested terrain highlight the need for dedicated UAV forest datasets to achieve operational reliability.
\end{itemize}

Taken together, these findings suggest that multimodal integration---combining lightweight fusion for per-frame enhancement with multi-view AOS for occlusion suppression---offers a promising path forward. The pipeline is not deployment-ready but establishes a baseline architecture that can guide future UAV perception systems.

\section{Conclusion}
This paper presented a proof-of-concept multimodal framework for detecting objects beneath sparse forest-canopy occlusion. LiDAR experiments showed limited utility for direct detection in the tested conditions, while visible--thermal fusion improved target saliency and AOS enhanced visibility in synthetic multi-view forest imagery. A YOLOv5 detector fine-tuned on thermal imagery achieved approximately $0.83$ mAP on the top three FLIR classes, indicating that lightweight thermal object detection is feasible. Together, these results establish an initial baseline for UAV-assisted search-and-rescue and surveillance under sparse canopy.  Future work should prioritize the following directions:
\begin{itemize}[leftmargin=*]
    \item Construction of a dedicated UAV-based visible--thermal forest dataset with annotated humans and controlled occlusion levels.
    \item Development of a real-time onboard pipeline integrating calibration, pose estimation, fusion, AOS-based refocusing, and detection.
    \item Evaluation of AOS under realistic UAV motion, imperfect pose estimates, and environmental variability.
    \item Exploration of learning-based fusion and transformer-based multimodal detectors once sufficient forest-specific data is available.
\end{itemize}

By combining lightweight fusion with multi-view geometry, and by grounding detection in dedicated forest datasets, future UAV systems can move toward reliable, certifiable perception under canopy occlusion.

\end{document}